\documentclass[10pt,twocolumn,letterpaper]{article}

\usepackage{cvpr}              
\usepackage{bbm}

\usepackage[dvipsnames]{xcolor}

\definecolor{cvprblue}{rgb}{0.21,0.49,0.74}
\usepackage[pagebackref,breaklinks,colorlinks,citecolor=cvprblue]{hyperref}

\def\paperID{330} 
\def\confName{3DV\xspace}
\def\confYear{2024\xspace}

\title{CoARF: Controllable 3D Artistic Style
Transfer for Radiance Fields}

\author{Deheng Zhang$^1$ \hspace{5mm}  Clara Fernandez-Labrador$^2$ \hspace{5mm}  Christopher Schroers$^2$\\
\\
$^1$ETH Zürich \hspace{5mm}  $^2$DisneyResearch$|$Studios\\
}

\begin{document}
\maketitle
\begin{abstract}
 Creating artistic 3D scenes can be time-consuming and requires specialized knowledge. To address this, recent works such as ARF \cite{zhang2022arf}, use a radiance field-based approach with style constraints to generate 3D scenes that resemble a style image provided by the user. However, these methods lack fine-grained control over the resulting scenes. In this paper, we introduce Controllable Artistic Radiance Fields (CoARF), a novel algorithm for controllable 3D scene stylization. CoARF enables style transfer for specified objects, compositional 3D style transfer and semantic-aware style transfer. We achieve controllability using segmentation masks with different label-dependent loss functions. We also propose a semantic-aware nearest neighbor matching algorithm to improve the style transfer quality. Our extensive experiments demonstrate that CoARF provides user-specified controllability of style transfer and superior style transfer quality with more precise feature matching.
\end{abstract}    
\begin{figure*}
    \centering
    \includegraphics[width=0.85\textwidth]{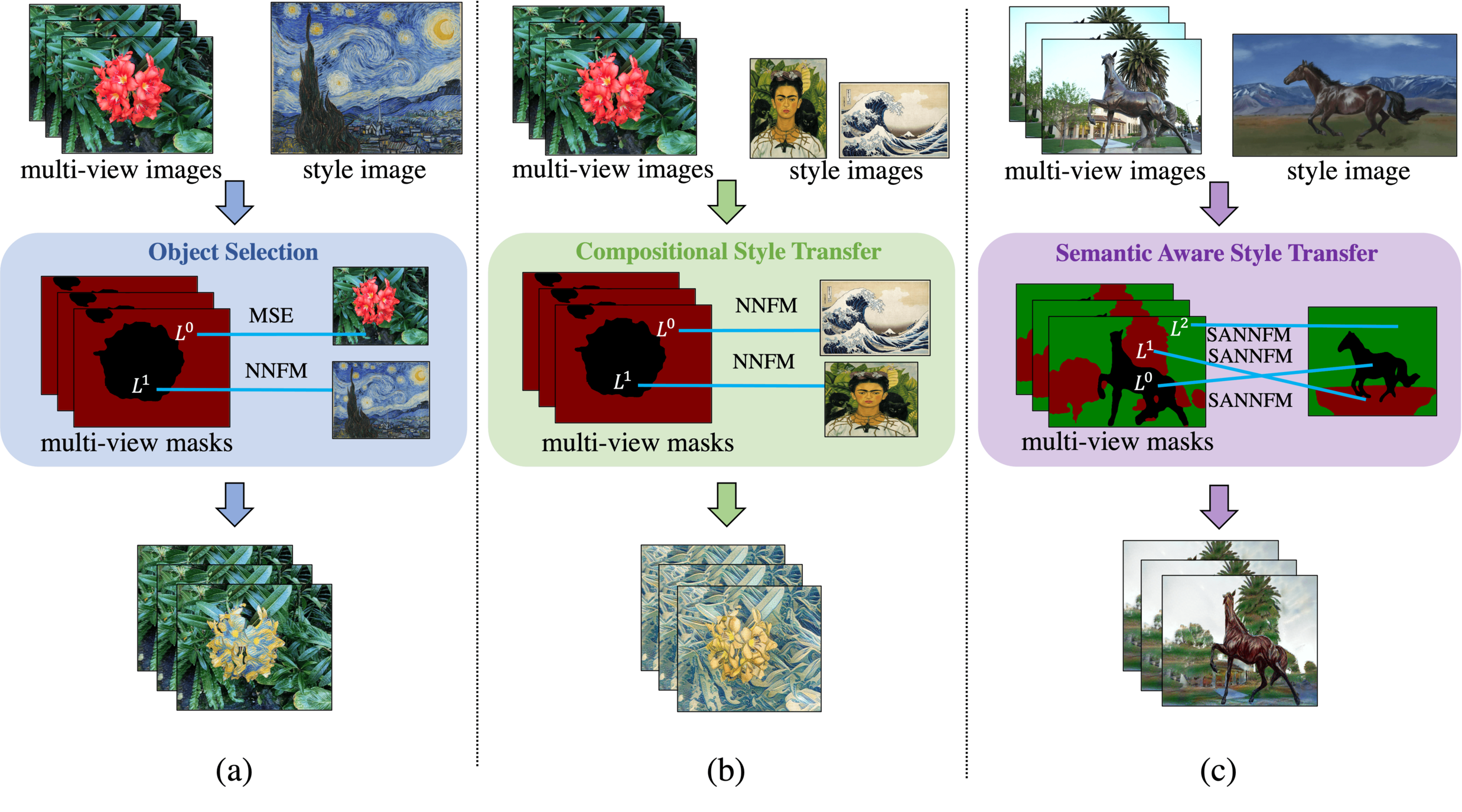}
    \caption[Our pipeline]{\textbf{CoARF Overview.} Given a set of multi-view ground truth images and style images, our controllable style transfer model allows the user to perform object selection \textup{(a)}, compositional style transfer \textup{(b)}, and semantic-aware style transfer \textup{(c)} by using 2D mask-based optimization with different spatial-dependent loss definitions to achieve stylization.%
      \label{fig:2Dmask}}
\end{figure*}
\section{Introduction}
\label{sec:intro}

\indent As the film and game industry continue to evolve, there is an increasing demand for more accessible methods to create and manipulate 3D scenes. Radiance field-based methods, such as Neural Radiance Fields (NeRF) \cite{mildenhall2021nerf}, gained widespread attention due to their ability to produce high-quality novel-view images without the requirement of 3D assets like meshes or point clouds. Instead, they use
implicit representation to encode the 3D scene information. However, despite the impressive photorealism achieved by these methods, the style richness of the resulting 3D scenes is still limited. To address this, researchers have turned their attention to style transfer for 3D content, which has gained significant interest as a means of providing inspiration and tools to 3D artists. There are many different style transfer approaches based on NeRF. For instance, \cite{chiang2022stylizing} use a separate HyperNetwork to control the input features based on arbitrary style images, while \cite{huang2022stylizednerf, li2023instant, liu2023stylerf} integrate AdaIN \cite{huang2017arbitrary} module to the NeRF pipeline to achieve style transfer. These methods cannot be easily integrated with NeRF variants such as \cite{fridovich2022plenoxels, muller2022instant} since the structure is tailored for style transfer. As a comparison, the Artistic Radiance Fields (ARF) \cite{zhang2022arf} applies a nearest neighbor feature matching (NNFM) style constraint to fine-tune the radiance information, thus can be generalized to NeRF variants and achieves superior stylization quality compared with the previous methods. However, artists need to control the style transfer target in practical applications, and ARF lacks fine-grained controllability in the stylization process. Other recent works \cite{liu2023stylerf, wang2022nerf} provide controllability for compositional style transfer and text-based style transfer, but the resulting quality is inferior and the semantic correspondence is not well captured during style transfer for these methods.

In this paper, we propose Controllable Artistic Radiance Fields (CoARF), which endows a baseline model \cite{zhang2022arf} with three controllability modules for 3D style transfer as shown in Figure \ref{fig:2Dmask}. Namely, CoARF enables object selection, compositional style transfer and semantic-aware style transfer using multi-view 2D mask-based optimization and masks defined by a user. Similar to the 2D case \cite{risser2017stable}, a naïve solution to achieve controllability is to define different optimization loss functions for different 3D regions. However, it is non-trivial to obtain the precise 3D object in practice. Therefore, we use multi-view 2D mask-based optimization with different loss definitions to achieve different controllable style transfer tasks and prove that multi-view optimization can automatically correct the error gradient.

In more detail, the object selection module allows the user to select which parts of the 3D scene should be stylized while keeping the rest photorealistic. To do so, we optimize the user's selected area using NNFM loss, and use Mean Squared Error (MSE) to preserve the rest photorealistic. The compositional style transfer module allows transferring of different styles to different parts of the 3D scene by using specific NNFM loss for corresponding mask labels. Last, our semantic-aware style transfer module uses 2D mask labels to match semantic regions between the 3D scene and the provided style image. We also show that while VGG features successfully encode texture, structure and color, they fail to encode semantic information. Therefore, we proposed a novel Semantic-aware Nearest Neighbor Feature Matching (SANNFM) algorithm which leverages a weighted sum of VGG feature distance and LSeg feature distance for stylization to improve general style transfer quality.

We summarize our contributions as follows: 1) We propose a multi-view 2D mask-based optimization framework to solve the general 3D controllable style transfer problem. 2) By leveraging the core framework, we can achieve object selection, compositional style transfer, and semantic-aware style transfer. 3) Our semantic-aware 3D style transfer algorithm utilizes a semantic-based nearest neighbor matching technique, which achieves better style transfer quality. Through extensive experimentation, we show that CoARF provides fine-grained control over the style transfer process and yields superior results compared to state-of-the-art algorithms \cite{zhang2022arf,liu2023stylerf}.

\section{Related Work}

\textbf{2D Style Transfer. } 2D style transfer can be achieved through various methods such as non-parametric algorithms \cite{kyprianidis2012state} or texture synthesis \cite{portilla2000parametric}. However, since \cite{gatys2016image} introduced the use of a convolutional neural network, there has been significant improvement in performance. This approach utilizes features extracted from the VGG-19 \cite{simonyan2014very} network to calculate a Gram matrix-based style constraint, and optimizes the output from a noise image. Follow-up methods \cite{li2016combining,risser2017stable, li2017demystifying, kolkin2019style,liao2017visual} use different content and style constraints to iteratively optimize the noise image. Recently, nearest neighbor-based algorithms \cite{li2016combining,liao2017visual,kolkin2022neural} have been proposed to match content and style features and minimize the cosine distance between them. ARF \cite{zhang2022arf} applies a similar nearest neighbor-based loss as \cite{kolkin2022neural}, despite the loss function being only applied to the final scale instead of augmenting the style images. In our work, we improve upon ARF by modifying the nearest neighbor-finding algorithm using semantic information, which results in better style transfer outcomes. There are also several efforts \cite{kim2022controllable, jing2018stroke} to explore the controllability of 2D style transfer from the level of stylization and stroke perspectives. Previous works \cite{risser2017stable,zhao2020automatic,huang2019style,huang2023composer} introduced user-defined masks or automatically generated soft masks for semantic-aware style transfer. Our work is similar to \cite{risser2017stable}, but we propose a novel semantic-aware method to improve feature matching.



\noindent \textbf{Radiance Fields. } NeRF \cite{mildenhall2021nerf} proposed a radiance field-based representation of the 3D scene. The method is trained using multi-view 2D images for which the camera pose is provided and uses MLPs to predict the density and radiance of a given sample on the ray. Then the volumetric rendering equation is applied to render the image for different views. To further enhance the quality and diversity of rendered scenes, several tasks have been explored, including NeRF rendering performance improvements \cite{barron2021mip,verbin2022ref,xu2022point}, NeRF speedup \cite{fridovich2022plenoxels,chen2022tensorf,muller2022instant}, dynamic scene rendering with radiance fields \cite{park2021nerfies,park2021hypernerf,pumarola2021d, peng2021animatable}, 3D scene editing and control \cite{lazova2023control, kania2022conerf, wang2022clip, kobayashi2022decomposing}, view consistent segmentation \cite{zhi2021place, fan2022nerf, wang2022dm, miao2023volumetric}, and novel view synthesis for real-world scenarios \cite{martin2021nerf,mildenhall2022nerf,xiangli2022bungeenerf}. In this work, we use Plenoxels \cite{fridovich2022plenoxels} as our backbone model for radiance prediction, which propose a sparse voxel model that matches NeRF's rendering performance while being much faster. Our algorithm can also be generalized to other differentiable radiance-field-based algorithms as demonstrated in ARF \cite{zhang2022arf}.

\noindent \textbf{NeRF-based Style Transfer. } There are various ways to enable stylized radiance prediction, including adding a tailored MLP \cite{chiang2022stylizing,huang2022stylizednerf,chen2022upst} for stylization or applying AdaIN transformation to the input features of the radiance MLP \cite{li2023instant}. However, these methods suffer from unstable results during training due to the use of a tailored stylization module. Moreover, they cannot be easily generalized to NeRF variants \cite{fridovich2022plenoxels,muller2022instant}, since the original scene encoded in NeRF is not edited.  In contrast, ARF \cite{zhang2022arf} applies an NNFM loss to the rendered image of the pre-trained radiance fields and fine-tunes the radiance information, resulting in exceptional stylization results compared to other methods. However, ARF lacks fine-grained controllability in the stylization process. An alternative approach, StyleRF \cite{liu2023stylerf}, uses volumetric rendering to integrate features in the scene and deferred style transformation to enable controllability for compositional style transfer. Nevertheless, the stylization quality is worse than ARF, because volumetric rendering for neural features is not physically-based. Our algorithm introduces fine-grained controllability to ARF using a 2D mask-based optimization and achieves better style transfer quality compared with existing methods. 

\noindent \textbf{Controllable NeRF. } The controllability of NeRF has been explored in various directions. For instance, \cite{lazova2023control} allows compositing objects from different scenes. \cite{kania2022conerf} enables dynamic facial expression control using 2D annotations, while  \cite{yuan2022nerf, garbin2022voltemorph, peng2022cagenerf, xu2022deforming, jambon2023nerfshop} edit the query for NeRF trained in static scenes, achieving user-defined deformation on the given scene. CLIPNeRF and SINE \cite{wang2022clip, bao2023sine} enable text-driven editing, whereas \cite{kobayashi2022decomposing} distills the 2D semantic feature from LSeg \cite{li2022language} to train 3D semantic feature using volumetric rendering, enabling editing including colorization, translation, deletion, and text-driven editing. In our project, we use LSeg \cite{li2022language} to provide a 2D mask for optimization and calculate semantic features to assist semantic-aware style transfer. However, we use 2D masks directly for controllability, unlike \cite{kobayashi2022decomposing}, which calculates explicit 3D label distribution for controllability.

\noindent \textbf{Language-driven Recognition. } The field of language-driven recognition models has gained a lot of attention after the release of CLIP \cite{radford2021learning}. CLIP utilizes text-image encoders to encode image and text features into the same space via contrastive pre-training, which enables zero-shot class prediction. LSeg \cite{li2022language} builds on this work by using the pre-trained text encoder from CLIP and an image encoder based on ViT \cite{dosovitskiy2020image} to achieve pixel-wise semantic feature prediction. Other works \cite{vinker2022clipasso,gal2022stylegan,michel2022text2mesh,mishra2022clip} have shown that the loss function defined by CLIP features has significant controllability for semantic information. In our work, we utilize a combination of semantic features from LSeg and features from VGG to guide nearest-neighbor searching and improve the robustness of style transfer.
\section{Background}
The general representation of the radiance fields is a mapping $f:\mathbb{R}^5 \mapsto \mathbb{R}^3$, which takes a 3D position $\textbf{x}$ and direction $\textbf{d}$ as input and outputs density $\sigma$ and radiance $\textbf{c}$:
\begin{equation}
\begin{aligned}
   \sigma, \textbf{c} = \text{RADIANCEFIELD}(\textbf{x}, \textbf{d}) .
\end{aligned}
\end{equation}
 The radiance for a given ray can be calculated as the following weighted average by uniform sampling points $\textbf{x}_i$ along the ray:
\begin{equation}
\begin{aligned}
   & \textbf{c}(\textbf{r}) = \sum^N_{i=1} w_i \textbf{c}_i, \;\; w_i=T_i(1 - exp(-\sigma_i\delta_i)),
\end{aligned}
 \label{eqn:vol_render}
\end{equation}
where $T_i = exp(-\sum^{i-1}_{j=1}\sigma_j\delta_j)$. The radiance fields function is optimized by the $L_2$ distance between the ground truth and the rendered image. In particular, we use Plenoxels \cite{fridovich2022plenoxels} as the radiance fields function, which represents the radiance as spherical harmonics functions in a voxel grid. Our CoARF is generalizable to the radiance fields representations.

In ARF \cite{zhang2022arf}, the radiance field is firstly pre-trained using the multi-view ground truth (content) images. Then, the density field is fixed, and ARF only optimizes the radiance prediction part of the radiance field. ARF fine-tunes the photorealistic radiance field by defining an averaged pixel-wise loss function on the rendered image:
\begin{equation}
\begin{aligned}
L = { 1 \over N} \sum_{x,y}( l_{\text{nnfm}}(\textbf{F}_r(x,y), \textbf{F}_s) &+ \lambda \cdot l_2(\textbf{F}_r(x,y), \textbf{F}_c(x,y)))\\
&+\lambda_{\text{tv}} \cdot l_{\text{tv}}, 
\end{aligned}
\end{equation}
\noindent where $N$ is the number of pixels, $\textbf{F}_c$, $\textbf{F}_s$, and $\textbf{F}_r$ are features extracted from the multi-view content (ground truth), style, and rendered images using VGG \cite{simonyan2014very} encoder. $\textbf{F}(x,y)$ denotes the feature vector at pixel location $(x,y)$. $l_2$ represents pixel-wise MSE and $l_{tv}$ represents total variation loss. The pixel-wise NNFM style loss is defined as the nearest cosine distance in the style image:
\begin{equation}
\begin{aligned}
  l_{\text{nnfm}}(\textbf{F}_r(x,y), \textbf{F}_s) &=  \min_{x', y'}D(\textbf{F}_r(x,y), \textbf{F}_s(x', y')) \\
  D(\textbf{v}_1, \textbf{v}_2) &= 1 - {\textbf{v}_1^T \textbf{v}_2 \over \sqrt{\textbf{v}_1^T \textbf{v}_1\textbf{v}_2^T \textbf{v}_2}} .
\end{aligned}
 \label{eqn:cosine}
\end{equation} 
ARF also applies color transfer to the voxel grid when they use Plenoxels as the backbone and the multi-view ground truth image before and after the stylization, which aligns the distribution of the RGB values to the style image.
\section{Methodology}
\begin{figure*}
    \centering
    \includegraphics[width=0.9\textwidth]{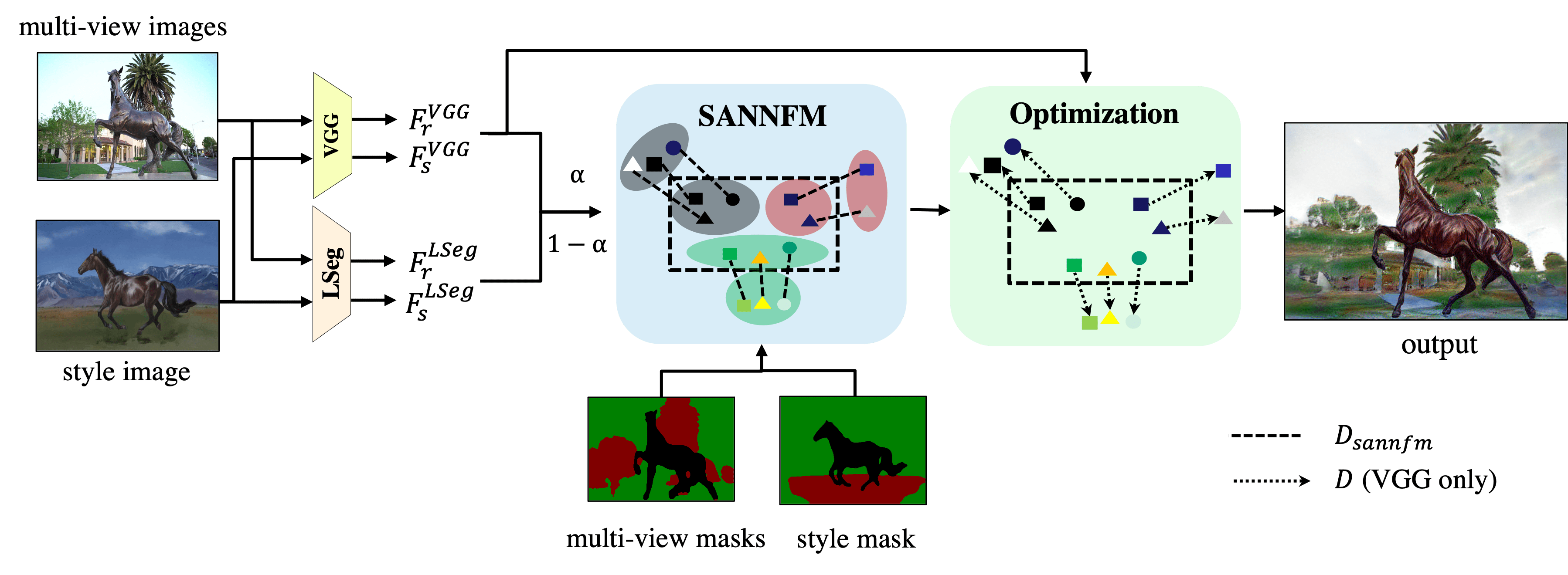}
    \caption[Pipeline of SANNFM]{\textbf{Pipeline of Semantic-aware Style Transfer.} The multi-view images and style image are used to extract VGG features $\textbf{F}_{r}^{VGG}, \textbf{F}_{s}^{VGG}$ and LSeg features $\textbf{F}_{r}^{LSeg}, \textbf{F}_{s}^{LSeg}$. Then the cosine distance is calculated and blended using the hyperparameter $\alpha$. As shown in the SANNFM module in the figure, we use the ellipses to represent different semantic labels, use the different shapes (square, triangle, circle) to represent semantic information of the pixels, use the color of the shape to represent the color and textural information of the pixels. The mixed distance is used to match the nearest neighbor in the style image with the same label for each pixel in the rendered image. Finally, the optimization uses VGG cosine distance only.
    	  
      \label{fig:sannfm}}
\end{figure*}

We propose a novel controllable and generalizable style transfer model for radiance fields, which not only enables object selection and compositional style transfer but also produces a better style transfer quality for semantic sensitive style images. We first introduce the core 2D mask-based algorithm to provide controllability for the radiance fields in Section \ref{sec:mask}. After that, we demonstrate how to apply this algorithm with different optimization constraints to achieve different controllable tasks in Section \ref{sec:diff_tasks}. Last, we provide implementation details of the controllable style transfer algorithms in Section \ref{sec:implementation}. 

\subsection{2D Mask-Based Optimization}
\label{sec:mask}

\begin{figure}
    \centering
    \setlength{\tabcolsep}{0.0130\linewidth}
    \begin{tabular}{@{}cc@{}}
    \includegraphics[width=0.387\linewidth]{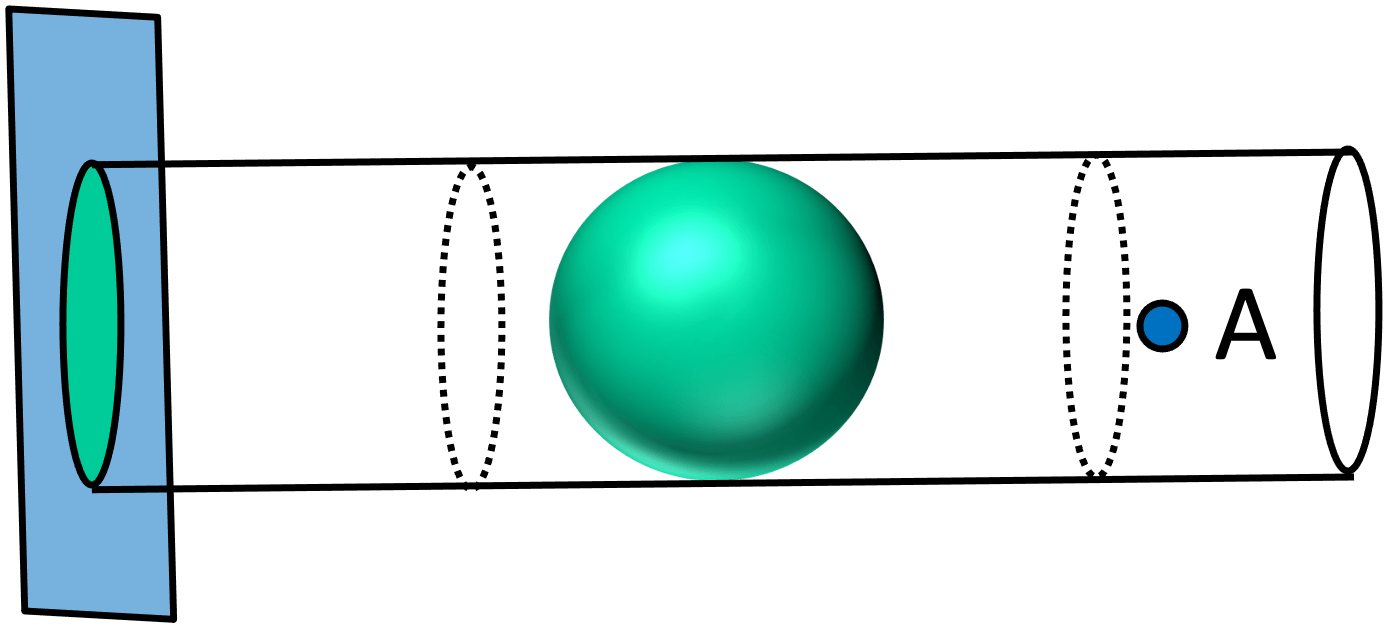}&
    \includegraphics[width=0.487\linewidth]{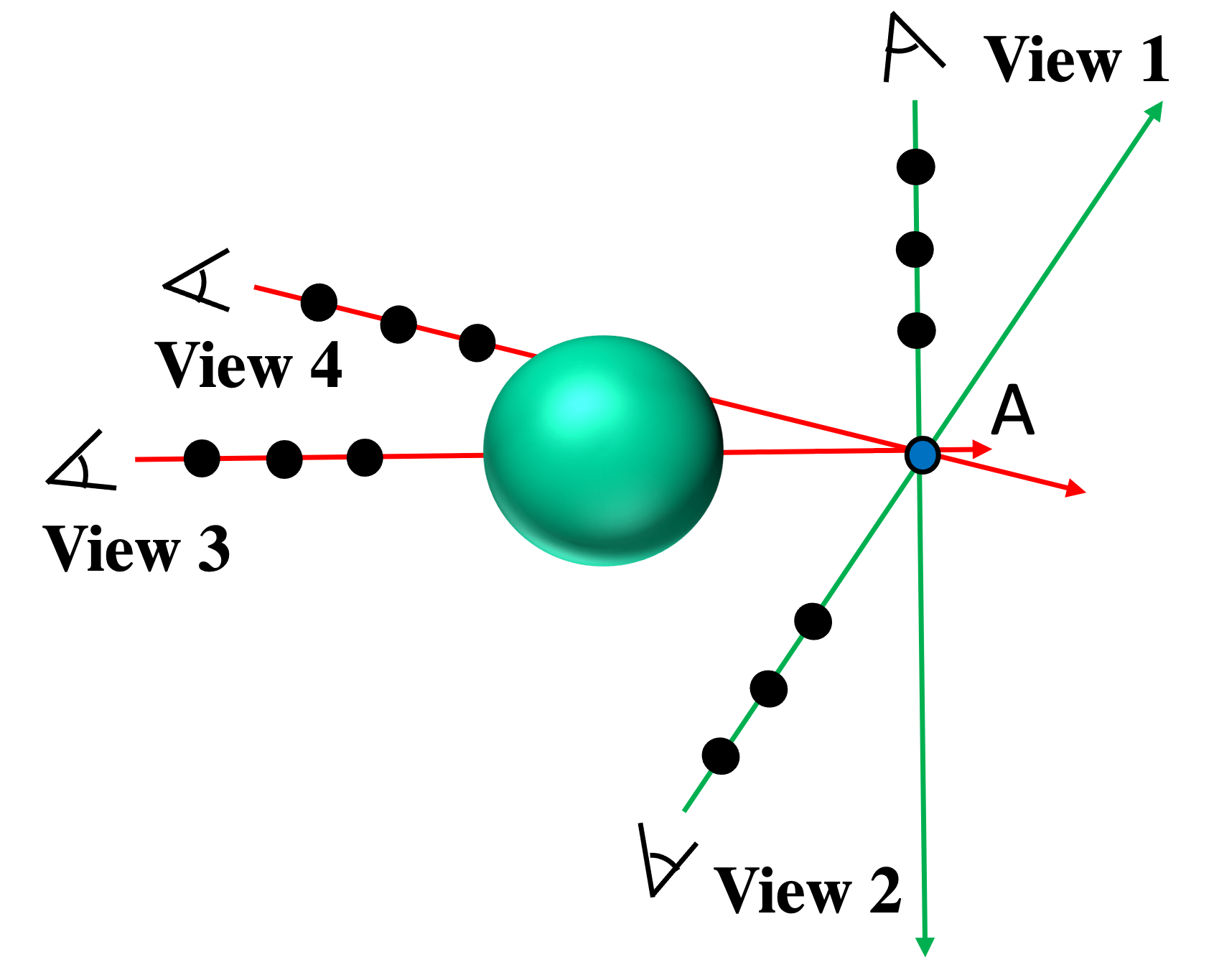}\\
    (a)&(b)\\
    \end{tabular}
    \caption[Multi-view Correction]{\textbf{Multi-view correction.}
    	  \textup{(a)} For one view optimization, point A should be optimized using background loss, but optimized using foreground object loss. 
		\textup{(b)} Multi-view ray passing through point A. The first two views have the correct label for point A, represented as green rays, last two views have an incorrect label for point A, represented as red rays. The final gradient is dominated by correct loss gradients in the first two views. 
      \label{fig:multi_view}}
\end{figure}

Given a rendered image from one camera view $\textbf{I}_r$, we can apply any 2D segmentation algorithm such as LSeg \cite{li2022language} to produce a segmentation mask $M_r$ with $M$ classes, such that each pixel $M_r(x,y) \in \{0, 1, ..., M\}$. We define the controllable loss function as:
\begin{equation}
\begin{aligned}
    L = ({1 \over N}\sum_{x,y} \sum_{m}\mathbbm{1}[M_r(x,y)=m] L^m(x,y)) +\lambda_{\text{tv}} \cdot l_{\text{tv}},
\end{aligned}
 \label{eqn:compose}
\end{equation}
where $L^m$ is a pixel-wise loss function defined for corresponding label $m$, and $\mathbbm{1}[\text{condition}]$ is an indicator function with output one if the condition holds and zero if the condition does not hold. In particular, we assign a specific loss function for each mask label. As shown in Figure \ref{fig:multi_view}(a), this loss function is problematic for one-view optimization because 2D mask-based loss optimizes a 3D column, instead of accurately optimizing the 3D object. As a result, point A would be optimized using the incorrect foreground loss function. However, as shown in Figure \ref{fig:multi_view}(b), with multi-view optimization, this error can be automatically corrected. According to the volume rendering equation, the final gradient for the radiance of point A $\nabla_{\textbf{c}_A}L$ is averaged among the multiple views:
\begin{equation}
\begin{aligned}
\nabla_{\textbf{c}_A}L = \sum_v w^v_A {\nabla L^{m_v}},
\end{aligned}
\end{equation}
where $w^v_A$ represents the contribution of point A during volume rendering in Equation \ref{eqn:vol_render} for view $v$, $m_v$ is the point A masking label from view $v$, and $\nabla L^{m_v}$ represents the gradient of the corresponding loss w.r.t the rendered pixel value. During optimization, A has the correct background label when visible and the wrong foreground label when occluded. According to the physical property of the volume rendering equation, since point A is visible from views 1 and 2, the weight becomes larger dominating the final gradient and optimizing the radiance with the correct background loss. The detailed proof can be found in the supplementary materials. Based on this optimization framework, we propose to combine different loss functions $L^m(x,y)$ based on the type of control we want to achieve.

\subsection{Controllable Optimization for Different Tasks}
\label{sec:diff_tasks}

\noindent \textbf{Object Selection. }
\label{sec:obj}
For object selection in 3D style transfer, we provide a 0-1 mask for each camera view to label objects that should be stylized (1) and objects that should be kept photorealistic (0). The loss function is the same as ARF for pixels with label 1, and we use MSE for pixels with label 0:


\begin{equation}
\begin{aligned}
L^m(x,y) = 
\begin{cases}
l_2 &m = 0\\
l_{\text{nnfm}} + \lambda \cdot l_2 &m = 1 \\
\end{cases}
.
\end{aligned}
\end{equation}
 Note that the preservation loss $L^{0}$ is a crucial constraint to correct the gradient as mentioned in Section \ref{sec:mask}. We show the effectiveness of $L^{0}$ in Figure \ref{fig:style_overflow}.
 
\noindent \textbf{Compositional Style Transfer. }
Compositional style transfer assigns different style images for different mask regions. The loss function for label $m$ in compositional style transfer is defined as:
\begin{equation}
\begin{aligned}
    L^m(x,y) =  l_{\text{nnfm}}(\textbf{F}_r(x,y), \textbf{F}_s^m) &+ \lambda \cdot l_2(\textbf{F}_r(x,y), \textbf{F}_c(x,y)),
\end{aligned}
\end{equation}
where the feature $F^m_s$ represents the feature map extracted from the style image with label $m$.

\noindent \textbf{Semantic-aware Style Transfer. } The original ARF uses VGG features to match the nearest neighbor in the style image. However, VGG features encode more textural, structural, and color information compared with semantic information. Therefore, we use LSeg  \cite{li2022language}, which is trained together with a text encoder to extract semantic information for the nearest neighbor matching. Our \textbf{S}emantic \textbf{A}ware \textbf{N}earest \textbf{N}eighbor \textbf{F}eature \textbf{M}atching (SANNFM) function performs nearest neighbor matching between content and style features in both VGG ($\textbf{F}_r^{VGG}, \textbf{F}_s^{VGG}$) and LSeg ($\textbf{F}_r^{LSeg}, \textbf{F}_s^{LSeg}$ ) spaces for each specific pixel $x,y$ with label $m$:
\begin{equation}
\begin{aligned}
 \text{SANNFM}(x,y, m) = \text{argmin}_{x', y' \in S}D_{\text{sannfm}} ,
\end{aligned}
\end{equation}
where $ S = \{x', y' | M_s(x',y') = m\}$, and the distance function $D_{\text{sannfm}}$ is defined as a weighted average of the VGG cosine distance and LSeg cosine distance:
\begin{equation}
\begin{aligned}
D_{\text{sannfm}} & = \alpha \cdot D(\textbf{F}^{VGG}_{r}(x,y), \textbf{F}^{VGG}_s(x',y')) \\
& + (1 - \alpha) \cdot D(\textbf{F}^{LSeg}_{r}(x,y), \textbf{F}^{LSeg}_s(x',y')),
\end{aligned}
\end{equation}
\noindent where $\alpha \in [0,1]$ is a hyperparameter to control the weight of VGG and LSeg features and $D(.)$ is the cosine distance defined in Equation \ref{eqn:cosine}. The SANNFM style loss for the pixel $x, y$ is defined as:
\begin{equation}
\begin{aligned}
 &  l_{\text{sannfm}}(m, \textbf{F}_r^{VGG}(x,y), \textbf{F}_s^{VGG}, \textbf{F}_r^{LSeg}(x,y), \textbf{F}_s^{LSeg}) \\
  & = {1 \over N} \sum_{x,y} D(\textbf{F}_{r}^{VGG}(x,y), \textbf{F}_{s}^{VGG}(\text{SANFFM}(x,y, m))).
\end{aligned}
\end{equation}
Intuitively, the pipeline is shown in Figure \ref{fig:sannfm}, we search the nearest neighbor for a content image pixel only in the corresponding mask label of the style image. And the nearest neighbor metric is defined as a combination of VGG cosine distance and LSeg cosine distance. Finally, we only optimize the VGG cosine distance between the two features, which means LSeg feature is only used to match the features. The loss
function for label $m$ in semantic-aware style transfer is defined as:
\begin{equation}
\begin{aligned}
  L^m(i, j)  &= l_{\text{sannfm}}(m, \textbf{F}_{r}^{VGG}(x,y), \textbf{F}_{s}^{VGG}, \textbf{F}_{r}^{LSeg}(x,y), \textbf{F}_{s}^{LSeg})
\\
& +  \lambda \cdot l_2(\textbf{F}_{r}^{VGG}(x,y), \textbf{F}_{c}^{VGG}(x,y)) .
\end{aligned}
\end{equation}

\subsection{Implementation Details}
\label{sec:implementation}
For the VGG features extraction, we follow the same approach as ARF, using the conv3 block of VGG-16. For the LSeg features, we use the LSeg encoder similar to \cite{kobayashi2022decomposing} to extract the feature for each pixel, and we use bilinear interpolation to match the resolution of VGG features. We set content loss weight $\lambda=0.001$ in forward-facing captures, and $\lambda=0.005$ in $360^\circ$ captures, and the smoothness weight $\lambda_{\text{tv}} = 1$. During colorization, different from ARF, we only apply color transfer to the ground truth images as preprocessing and do not change the voxel grid, since we observe that the color transfer on the whole feature grid has an inaccurate rendering result since it is not semantically aware. For object selection, we only apply color transfer to the selected object in the ground truth image. And for compositional style transfer and semantic-aware style transfer, we apply color transfer to different ground truth regions using the corresponding semantic parts. All experiments are conducted with a single NVIDIA RTX 3090 Ti GPU. 
\section{Experiments}
We conduct extensive experiments to evaluate object selection in Section \ref{sec:obj_exp}, compositional style transfer in Section \ref{sec:multi_exp}, and semantic-aware style transfer in Section \ref{sec:sem_exp}. We also validate our 2D mask-based optimization and SANNFM module in the ablation studies in Section \ref{sec:ablation}.

\noindent \textbf{Dataset. } We use a total of seven scenes including three forward-facing captures: \textit{Flower, Fortress, Horns}, from \cite{mildenhall2021nerf}, and four $360^\circ$ captures: \textit{Family, Horse, M60, Truck} from the \textit{Tanks and Temples} dataset \cite{knapitsch2017tanks}. We use style images from the ARF style dataset, cartoon movies and other paintings. 

\subsection{Object Selection}
\label{sec:obj_exp}
\begin{figure}
    \centering
    \setlength{\tabcolsep}{0.0130\linewidth}
    \includegraphics[width=1\linewidth]{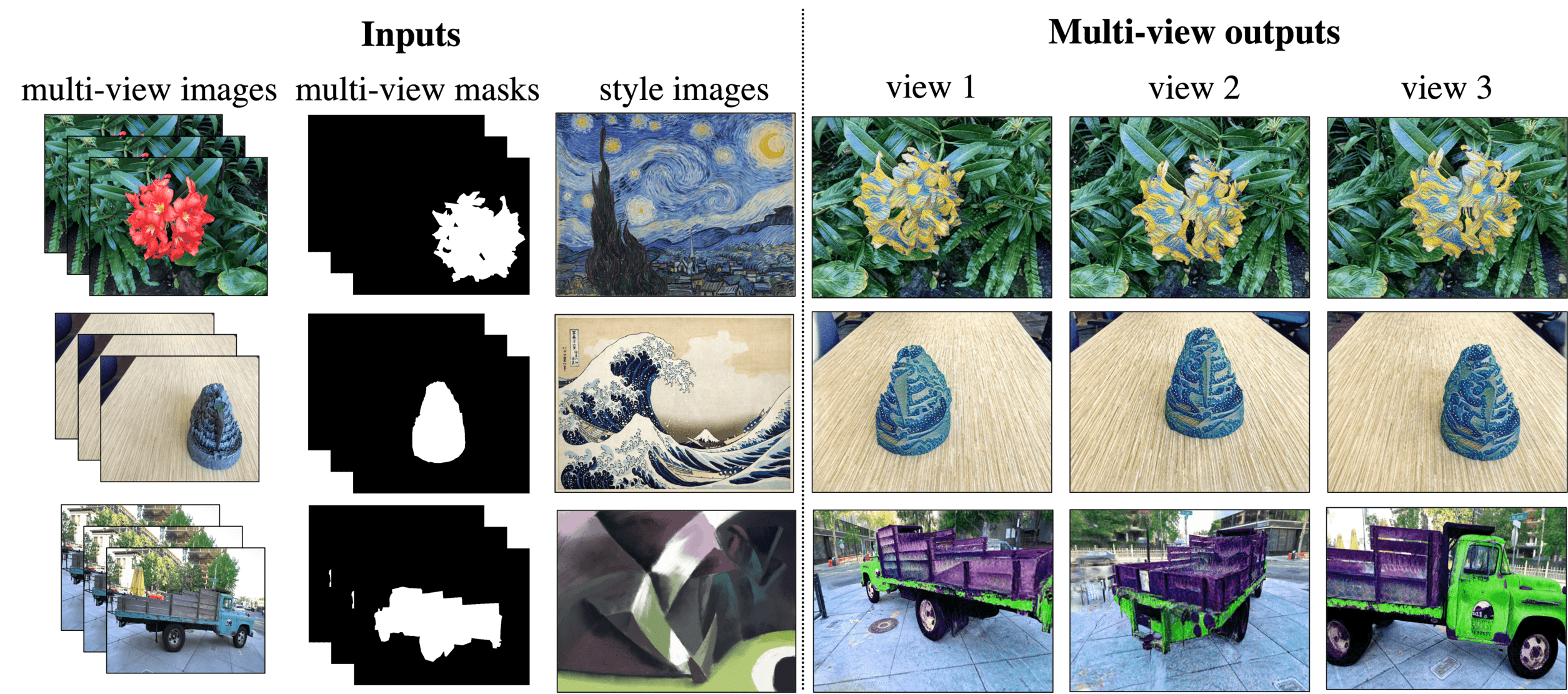}
    \caption[Object Selection Result]{\textbf{Object selection result.} Our algorithm can generate the stylized object for the selected region and keep other objects photorealistic. %
      \label{fig:obj_exp}}
\end{figure}

Our result for object selection is shown in Figure \ref{fig:obj_exp}. We use ground truth mask from \cite{fan2022nerf} and show the multi-view style transfer result for: \textit{Flower, Fortress}, and \textit{Truck} scene. As shown in the result, our method can correctly apply style transfer to the selected region, and leave other regions unchanged. 
\subsection{Compositional Style Transfer}
\label{sec:multi_exp}
\begin{figure}
    \centering
    \setlength{\tabcolsep}{0.0130\linewidth}
    \includegraphics[width=0.9\linewidth]{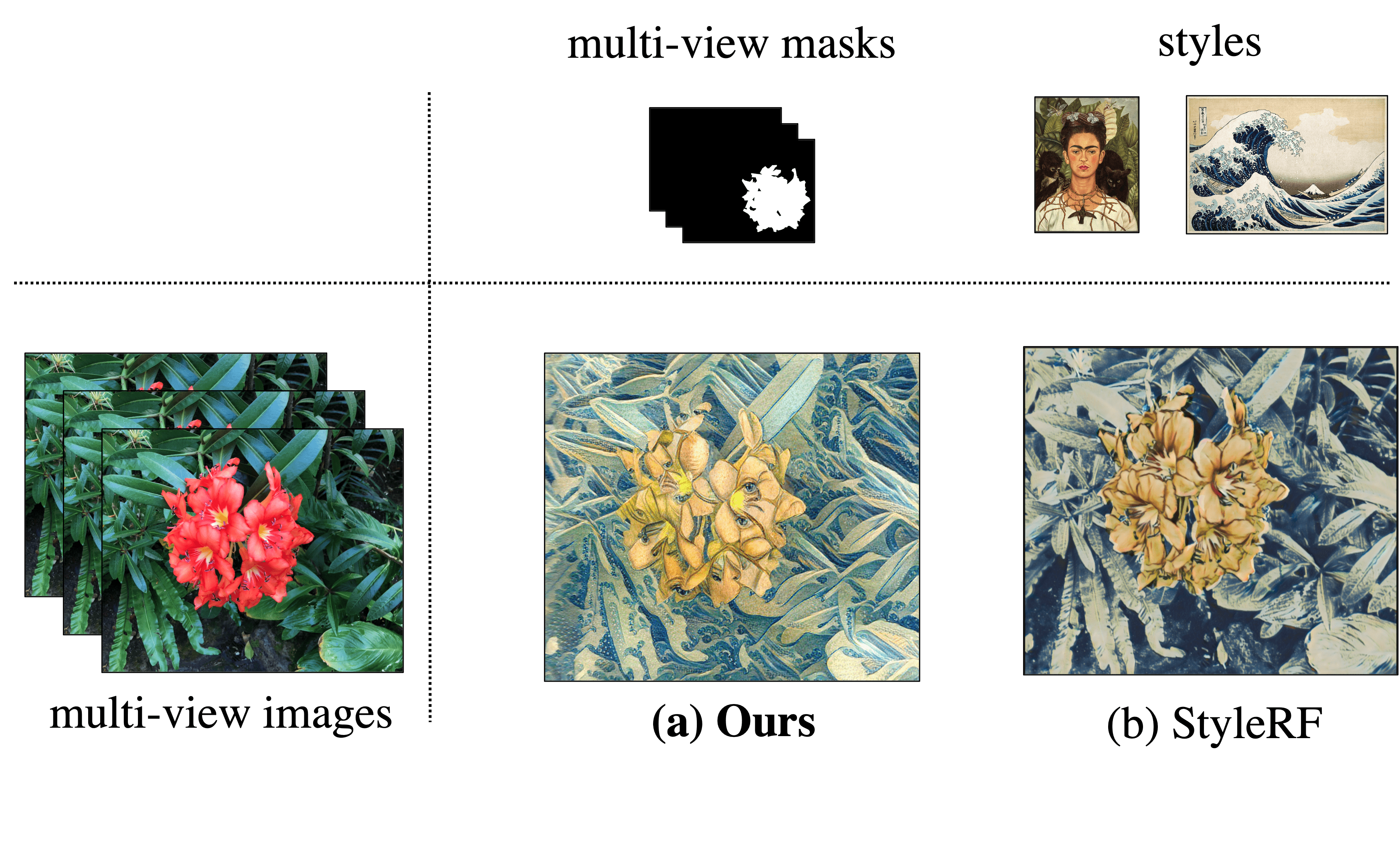}
    \caption[Qualitative Comparison of Compositional Style Transfer]{\textbf{Qualitative comparison of compositional style transfer.} Our result (a) can match the style image better in terms of the brushstrokes and color compared with StyleRF \cite{liu2023stylerf} (b). 
      \label{fig:compose_exp}}
\end{figure}

Compositional style transfer results are shown in Figure \ref{fig:compose_exp}. We compare our results with StyleRF \cite{liu2023stylerf}, which uses volume rendering to generate a 2D feature map, and applies style transfer after the rendering. Thanks to the robustness of NNFM style constraint, our method can generate the composed style scenes that match the style images better in terms of brushstrokes and color, while StyleRF \cite{liu2023stylerf}  generates over-photorealistic geometric structure, and fail to transfer the color of the style image. Also, we observe that our algorithm converges faster compared with StyleRF thanks to the efficient implementation of Plenoxels\cite{fridovich2022plenoxels}. Since our method is optimization based, it can be further generalized to other radiance fields \cite{barron2021mip,muller2022instant} to either improve rendering quality or fasten the stylization convergence. But StyleRF cannot readily take advantage of other radiance fields.
\subsection{Semantic-aware Style Transfer}
\label{sec:sem_exp}
\begin{figure*}
    \centering
    \includegraphics[width=0.87\textwidth]{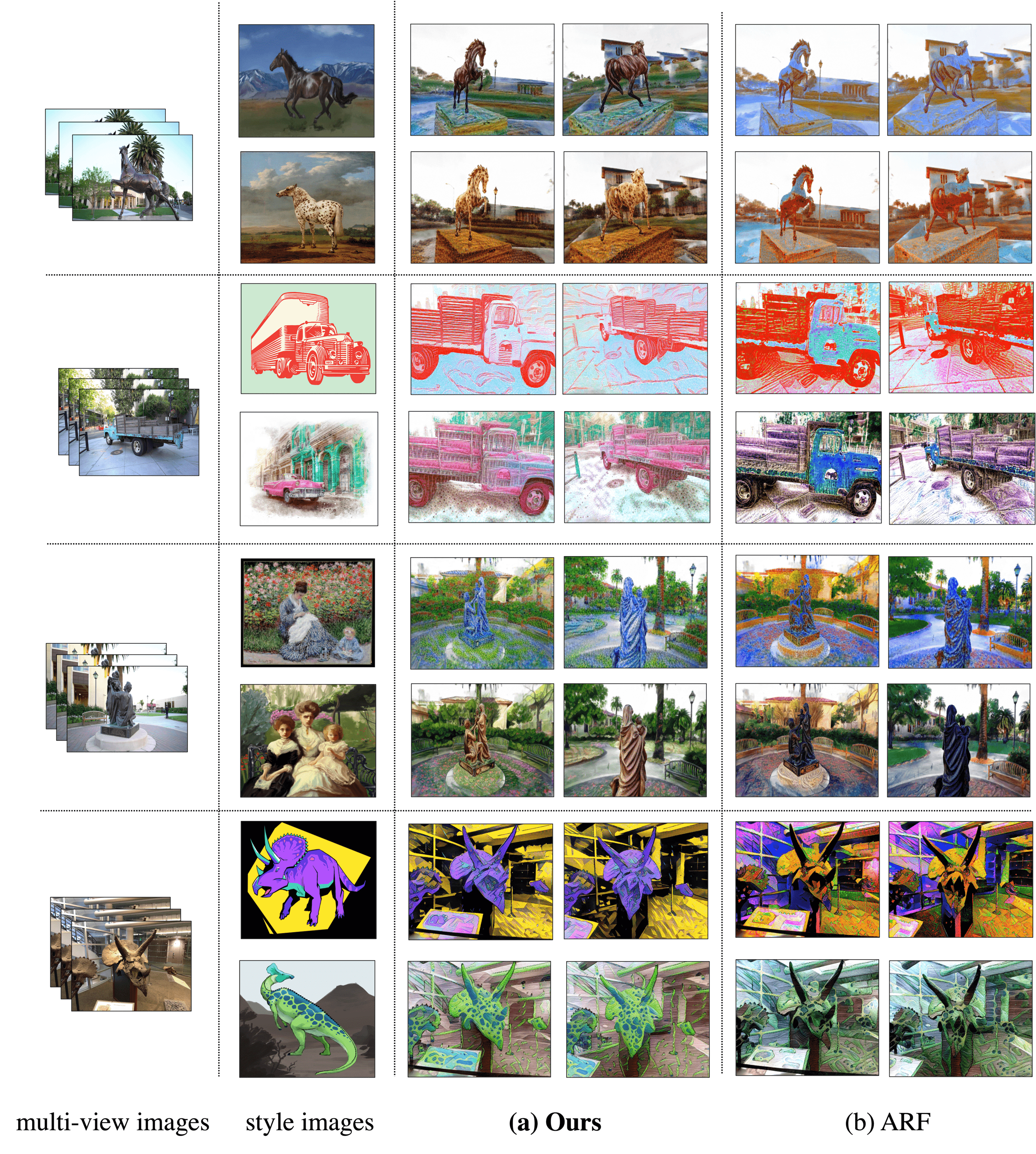}
    
    \caption[Comparison with the baseline method ARF\cite{zhang2022arf}]{\textbf{Comparison with the baseline method ARF\cite{zhang2022arf} on real-world forward-facing and \textit{Tanks and Temples} data.} Our method matches the semantics more faithfully compared with ARF. 
      \label{fig:compare}}
\end{figure*}

We show comparisons between our semantic-aware style transfer and ARF \cite{zhang2022arf} in Figure \ref{fig:compare}. We can observe that our results can match semantic information better compared with ARF. For example, in the horse scene, our results correctly match the color and texture of the horse (brown in the first style, dot in the second style), while our baseline ARF \cite{zhang2022arf} can only matches the green color in the style image since the horse in ground truth image is green. This is because we use the semantic mask with the LSeg features to guide the nearest neighbor matching, while ARF \cite{zhang2022arf} uses the VGG features only, which encodes more color and texture information.

\noindent \textbf{User Study. } We also conduct a user study to compare the style transfer quality between our semantic-aware style transfer and our baseline ARF \cite{zhang2022arf}. Similar to Figure \ref{fig:compare}, we compare fourteen stylization results using the seven scenes of our dataset and two style images per scene that share similar semantics. For each example, we provide sample images from the photorealistic scene, the style image and the result multi-view videos from our method and ARF, and we let a large group of users select a preferred stylization or mark both results as similar. As a result, users prefer our result over ARF \cite{zhang2022arf} $84\%$ of the time, and $11.1\%$ of the time think the results have similar quality. These results show our method is preferable most of the time.  We provide more detail in the supplementary materials.

\subsection{Ablation Studies}
\label{sec:ablation}
\begin{figure}
    \centering
    \setlength{\tabcolsep}{0.0130\linewidth}
    \includegraphics[width=0.8\linewidth]{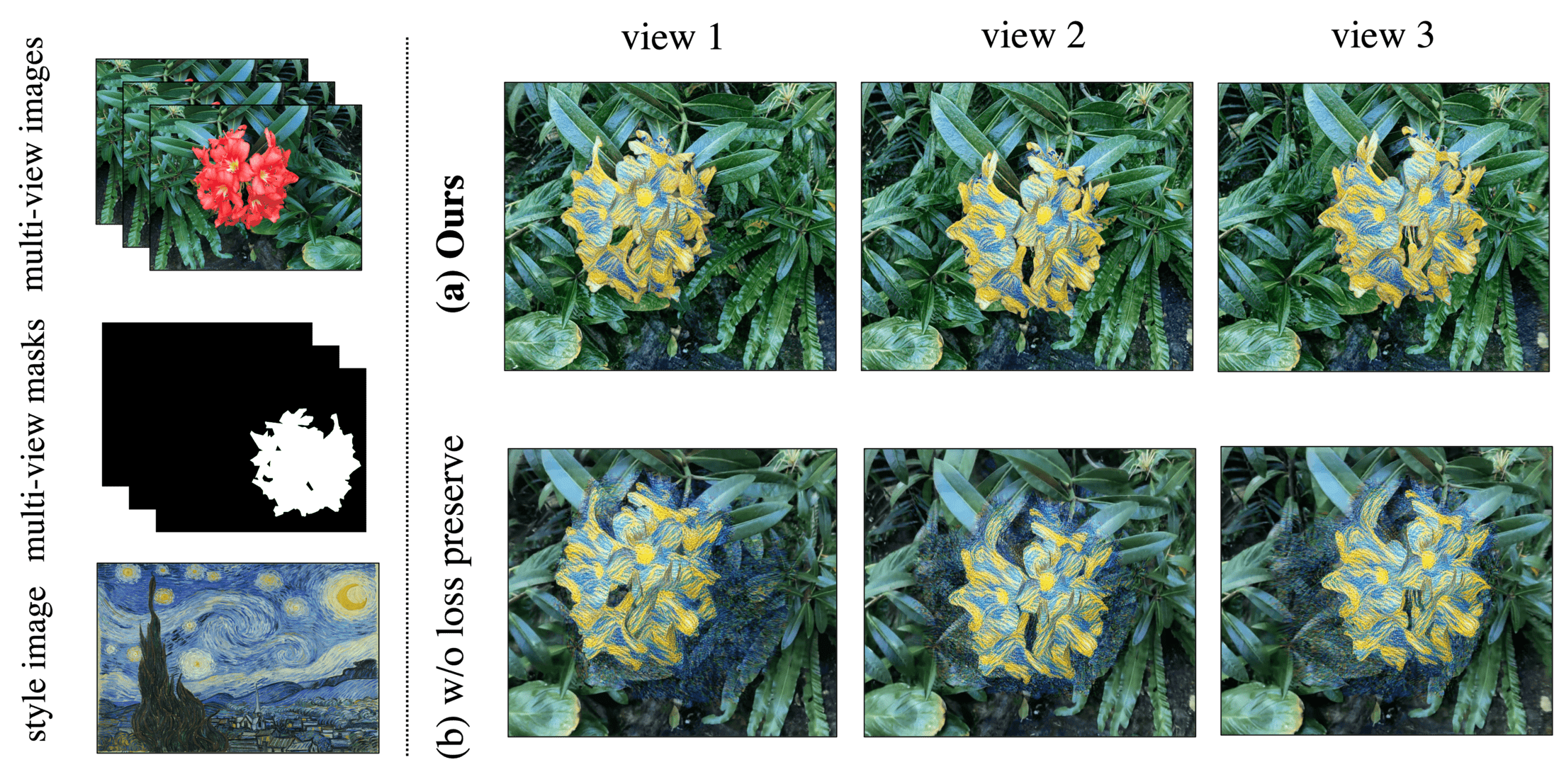}
    \caption[Validation of the Preservation Constraint]{\textbf{Validation of the preservation constraint.} \textup{(a)} shows our result, while (b) shows the result without the constraint. We can find that the preservation loss corrects the gradient, while style overflow happens when we remove this loss.%
      \label{fig:style_overflow}}
\end{figure}

\noindent \textbf{Multi-view Gradient Correction. } As mentioned in Section \ref{sec:mask}, the perservation constraint for object selection can correct the gradient for occluded objects. To prove the effectiveness of the constraint, we conduct an experiment to compare the stylized output with and without constraint. As shown in Figure \ref{fig:style_overflow}, without the preserving constraint, many voxels are optimized incorrectly. As a result, there are some noisy pixels nearby the selected object. We call this effect style overflow. With the constraint, the incorrect gradients are corrected by the constraint gradient, thus the object selection is cleaner. This experiment shows that multi-view optimization can automatically correct the error gradient due to the column extended from the 2D mask. 

\begin{figure}
    \centering
    \setlength{\tabcolsep}{0.0130\linewidth}
    \includegraphics[width=0.8\linewidth]{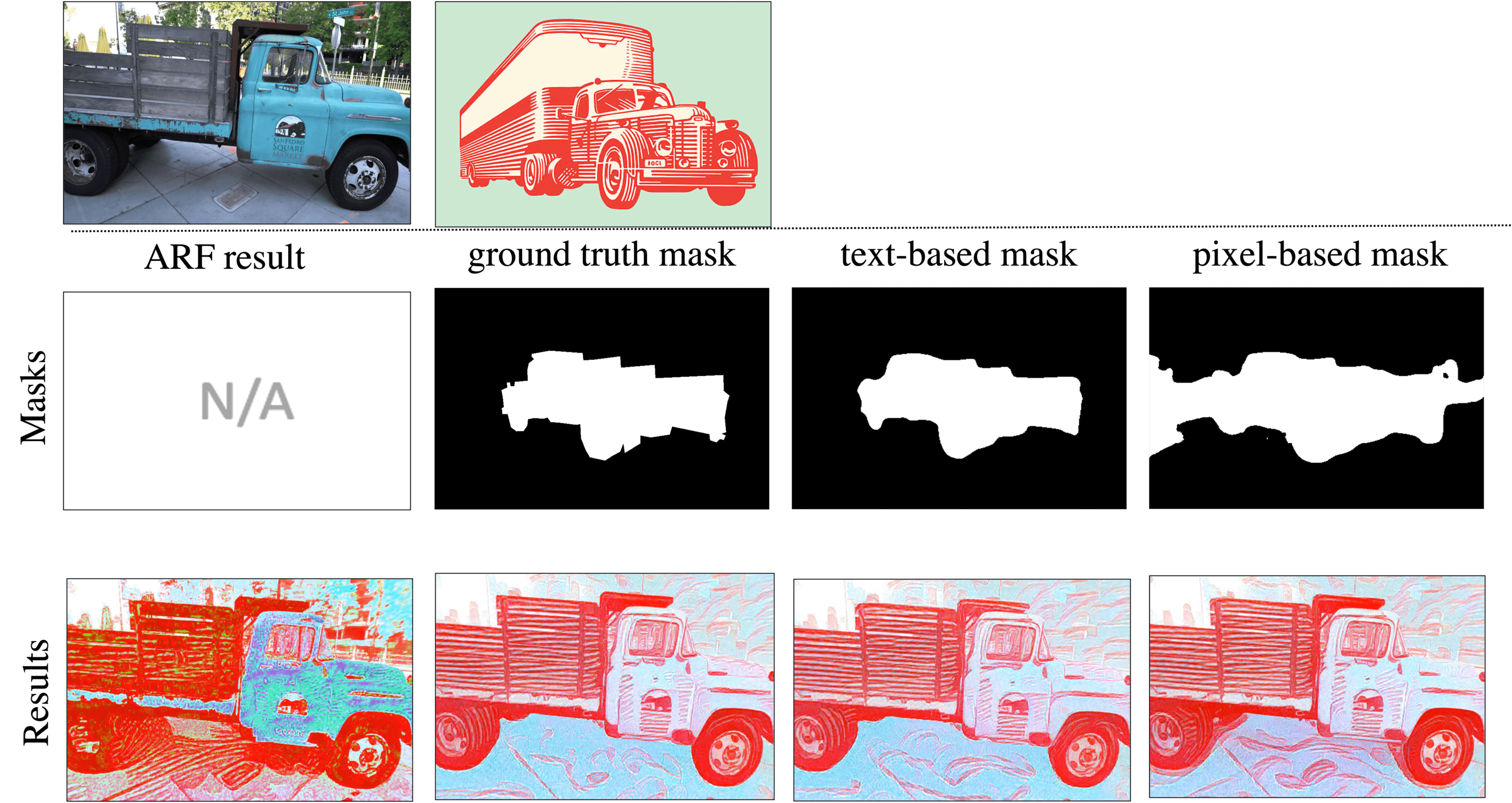}
    \caption[Effect of Mask Quality on the Style Transfer]{\textbf{Effect of mask quality on the style transfer.} We show how robust our method is with different 2D mask qualities. All of our methods generate more semantic-aware results compared with ARF \cite{zhang2022arf}. And the quality of style transfer does not drop so much when the view consistency is not guaranteed.
    	   . 
      \label{fig:mask_quality}}
\end{figure}

\noindent \textbf{Mask Quality. } It is difficult to obtain ground truth masks in practice, therefore, we use two methods to extract masks. We can either use a text (such as \textit{horse, flower}) or select a pixel belonging to an object directly in the image to get the mask using LSeg \cite{li2022language}. Thanks to our robust 2D mask-based optimization, we can use either ground truth masks or relatively more noisy masks generated by the aforementioned two methods. As shown in Figure \ref{fig:mask_quality}, although the pixel-based mask is noisy, our algorithm generates a reasonable result, which better matches the semantics in the style image significantly compared with ARF \cite{zhang2022arf}. And the result using the text-based mask has comparable quality compared with the ground truth mask result.

\begin{figure}
    \centering
    \setlength{\tabcolsep}{0.0130\linewidth}
    \includegraphics[width=0.8\linewidth]{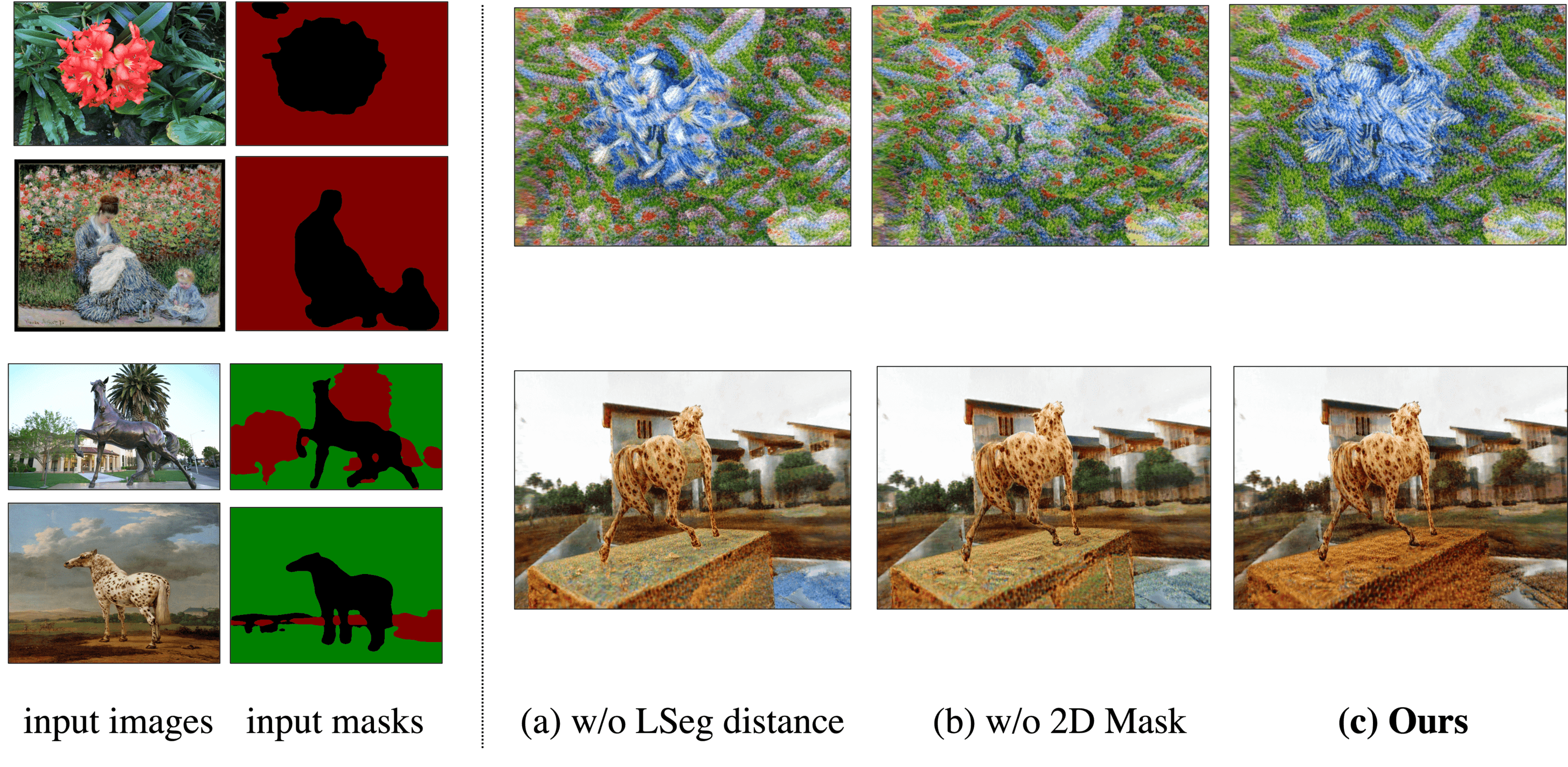}
    \caption[Ablation Studies for SANNFM]{\textbf{Ablation studies for SANNFM.}  The left column shows the ground truth images and masks, style images and masks. The right column shows (a) the result without LSeg distance (b) the result without mask constraint (c) our whole pipeline including LSeg distance and mask constraint. 
      \label{fig:sannfm_ablation}}
\end{figure}

\noindent \textbf{SANNFM. } Original ARF \cite{zhang2022arf} uses the cosine distance of the VGG features to match the nearest neighbor, while our method only matches the nearest neighbor with the same semantic label, and uses a combination of VGG and LSeg cosine distance. To prove the effectiveness of these two modifications, we perform an ablation study. As shown in Figure \ref{fig:sannfm_ablation}, in column a), if we only use mask constraint but no Lseg distance to find the nearest neighbor, we have some artefacts on the horse body in the stylized result. In column b) if we remove the mask and there is no semantic similarity between the ground truth and style image (flower scene), the model behaves as standard methods, whereas if there is semantic similarity (horse scene), the model successfully transfers the style taking the semantics into account. This shows that our algorithm is robust when there is a correspondent object in the style image, but the 2D mask is necessary to provide more user control options when there is no correspondence. Column c) shows our results using both LSeg distance and 2D masks, which successfully transfer the style in both examples. In general, our method with mixed distance metric and masked constraint has a more robust controllability for semantic-aware style transfer. 
\begin{figure}
    \centering
    \setlength{\tabcolsep}{\linewidth}
    \includegraphics[width=0.987\linewidth]{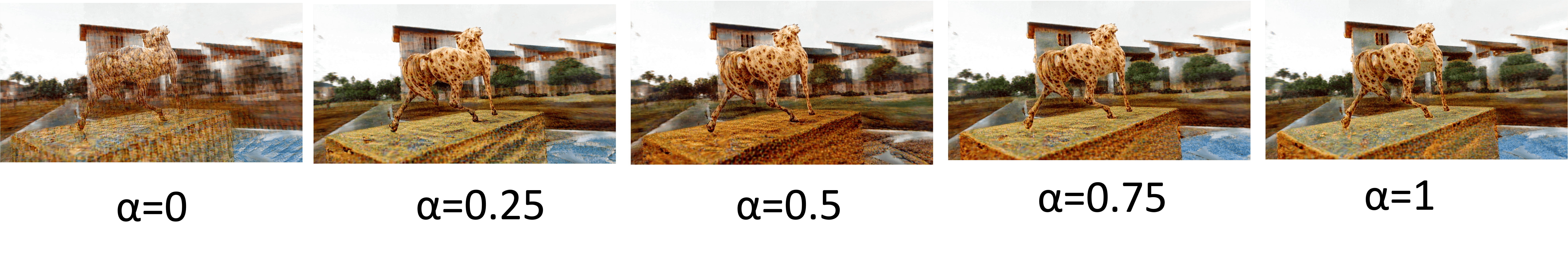}
    \caption[Different Combination of VGG and LSeg Features]{\textbf{Different combination of VGG and LSeg features.}  $\alpha = 0$ means we only use LSeg features, and $\alpha=1$ means we only use VGG features. Note that all results are using the mask constraint.
      \label{fig:vgg_lseg}}
\end{figure}

\noindent \textbf{VGG \& LSeg Features. } As shown in Figure \ref{fig:vgg_lseg}, we show style transfer results using a different combination of VGG and LSeg distance. We can observe that if we only use the VGG features, the features matching has some artefacts on the horse. But if we only use the LSeg features, the result becomes noisy. Therefore, a combination of VGG and LSeg distance gives the best result by balancing the semantic and textural information.

\section{Limitations}
In this section, we discuss some limitations of the presented method and potential improvements. We notice that large scale differences between similar semantic objects in the scene and the style images could lead to undesired results. For example, if the scene object has a larger scale than the style object, the algorithm tends to copy multiple style objects to fill the scene object. One potential solution could be to use the corresponding segmentation masks to adjust the resolution of the style image. Besides, following ARF, we freeze the density field and only optimize the radiance field. However, we believe that optimizing the density could produce richer stylization results. Furthermore, our semantic-aware style transfer model is slower than ARF due to the computational requirement of the LSeg module. 
\section{Conclusion}
In this paper, we propose different types of controllability including object selection, compositional style transfer, and semantic-aware style transfer. We achieve these tasks by multi-view 2D mask-based optimization with different constraints. This can be generalized to other optimization-based tasks such as CLIPNeRF \cite{wang2022clip} to achieve user-defined controllability. In addition,  we propose SANNFM that combines the mask constraint with a mixed distance of VGG features and LSeg features. Our experiments show that our approach not only has better quality and robustness but also has better controllability given the user-defined semantics.

\section{Acknowledgment}
Thanks to the artists Violaine Fayolle and Doriano vanEssen for their support of great artworks. 
{
    \small
    \bibliographystyle{ieeenat_fullname}
    \bibliography{main}
}
\clearpage
\setcounter{page}{1}
\maketitlesupplementary
\section{Derivation of the validity of 2D mask-based optimization}
\label{sec:proof}
Given a pixel, the RGB value of the pixel $\textbf{c}$ is calculated using a volumetric rendering equation:
\begin{equation}
\begin{aligned}
   \textbf{c} = \sum^N_{i=1} w_i \textbf{c}_i, \text{where}\ w_i=T_i(1 - exp(-\sigma_i\delta_i)), 
\end{aligned}
\end{equation}
and $T_i = exp(-\sum^{i-1}_{j=1}\sigma_j\delta_j)$. Note that each $i$ corresponds to a sample during ray marching. Then, we define the 2D mask-based controllable loss function for pixel $x, y$:
\begin{equation}
\begin{aligned}
    L(x, y) =  \sum_{m=0}^{M}\mathbbm{1}[M_r(x, y)=m] L^m(x, y),
\end{aligned}
 \label{eqn:compose}
\end{equation}
where $M_r(x, y)$ represents the masking label of the pixel. Therefore, we can calculate the gradient of the radiance of a sample $i$ using 2D mask-based optimization:
\begin{equation}
\begin{aligned}
    {\nabla_{\textbf{c}_i} L} &= 
    \sum_{m=0}^{M}\mathbbm{1}[M_r(x,y)=m] \nabla_{\textbf{c}_i}L^m(x, y) \\
    &= \nabla_{\textbf{c}_i}L^{M_r(x,y)}(x, y)\\ 
    &=w_i \nabla_{\textbf{c}}L^{M_r(x,y)}(x, y),
\end{aligned}
\end{equation}
where $\nabla_{\textbf{c}_i}L$ represents the gradient w.r.t the $i$-th sample radiance, and $\nabla_{\textbf{c}}L$ represents the gradient w.r.t the rendered pixel value. As we discuss in Section \ref{sec:mask} the transparent and the occluded part will be optimized incorrectly. But if we incorporate multi-view optimization, the gradient $\nabla_{\textbf{c}_i}L$ can be represented as:
\begin{equation}
\begin{aligned}
\nabla_{\textbf{c}_i}L = \sum_v w^v_i {\nabla_{\textbf{c}} L^{m_v}}, \\
\end{aligned}
\end{equation}
where $m_v$ represents the masking label of the rendered pixel in view $v$. Given an input view $v$, the weighting factor $w_i^v$ for transparent and occluded samples is relatively small, whereas for  visible samples is larger. As a consequence, the final gradient is dominated by the view in which sample $i$ is visible, being more relevant to the loss function and naturally leading to a correct optimization. Therefore, similar to the idea that the hash collision is avoided by the dominant samples in \cite{muller2022instant}, the update vector is dominated by visible samples.

\section{User Study}
As discussed in Section \ref{sec:ablation} of the paper, we observe that CoARF outperforms the ARF baseline by a large margin on average. However, there are also some stylization results that present more divided opinions. The reason is that the semantic subject in the style image has a similar color distribution to the semantic subject in the scene, thus VGG and LSeg feature has similar feature-matching performance.


\section{Mask Generation Method} 
\begin{figure}
    \centering
    \setlength{\tabcolsep}{0.0130\linewidth}
    \includegraphics[width=0.987\linewidth]{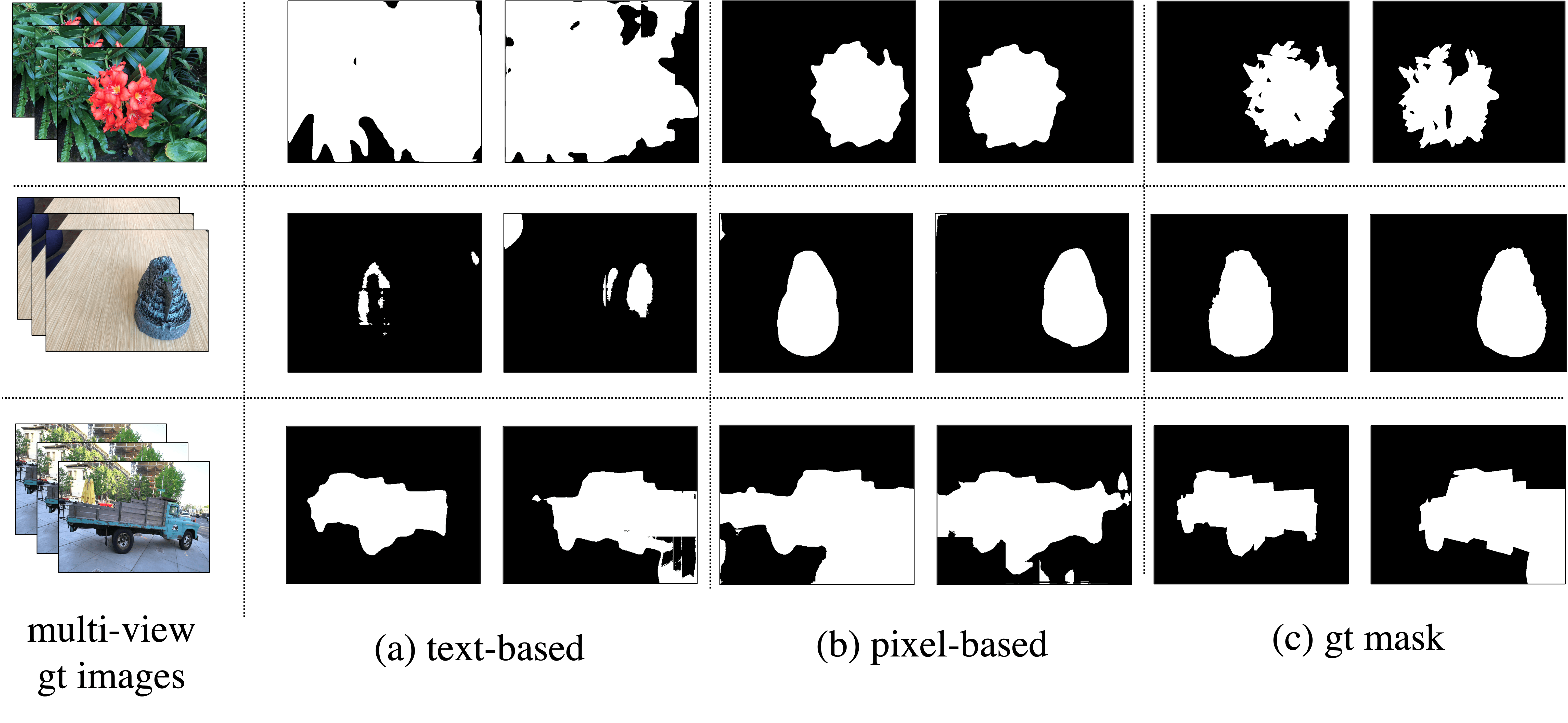}
    \caption[Comparison of Different Mask Generation Methods]{\textbf{Comparison of different mask generation methods.} The left column is the multi-view ground truth images. Images in column (a) are the masks generated by the text-based method as in \cite{li2022language}. Images in column (b) are the masks generated by the pixel-based method. Images in column (c) show the ground truth mask labelled manually. %
      \label{fig:mask_method}}
\end{figure}

\begin{figure}
    \centering
    \setlength{\tabcolsep}{0.0130\linewidth}
    \includegraphics[width=0.987\linewidth]{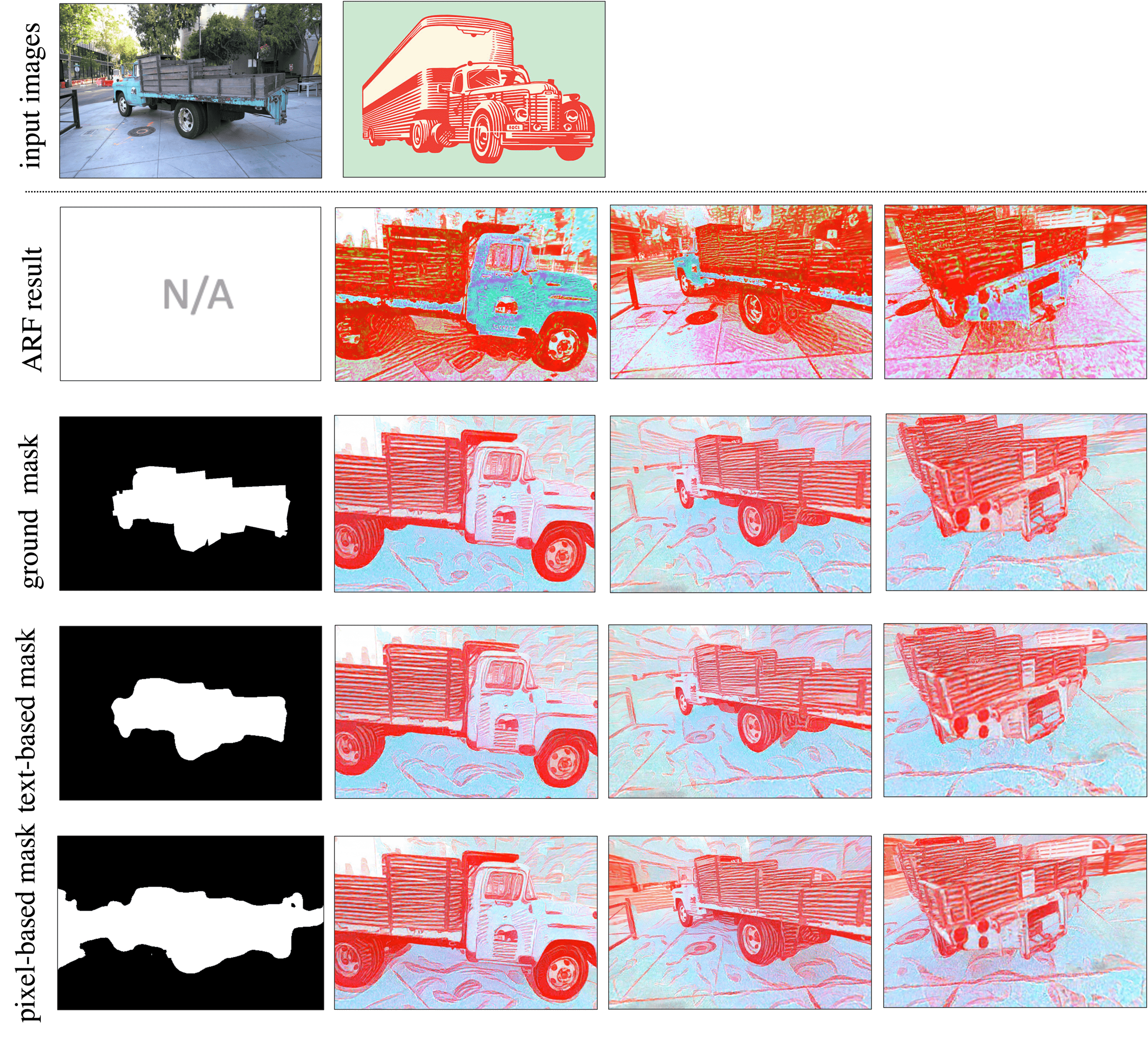}
    \caption[Effect of Mask Quality on the Style Transfer in Multiple Views]{\textbf{Effect of Mask Quality on the Style Transfer in Multiple Views.} This is a supplementary figure corresponding to Figure \ref{fig:mask_quality} of the paper.
    	   . 
      \label{fig:mask_mul}}
\end{figure}

Although ground truth masks have better quality and guarantee view consistency, it is difficult and time-consuming to manually label multiple view images in a consistent way. In our pipeline, we provide two methods to extract 2D masks using LSeg \cite{li2022language} image and text encoders. The first method follows the original LSeg idea. Given a text query provided by the user, LSeg embeds text and image pixels into a common space and generates masks by finding the minimum distance between the two embeddings.  The text query should specify the semantic similarities between content and style images to perform stylization. We propose a second method where the user simply selects a pixel in the image to mark the object and the mask is extracted based only on per-pixel embeddings similarity instead. The second method is especially desired when there are ambiguities or when the object has no clear definition. As shown in Figure \ref{fig:mask_method}, we can observe that for the flower scene text-based method fails to extract the mask. Because in the scene, the word "flower" also includes the leaves part semantically. And for the fortress scene, since "fortress" is not a common word, the performance of the text-based method is also worse than the pixel-based method. However, for the truck scene, since the truck object has many different components, the pixel-based method fails to capture the semantic information. It is necessary that we provide both methods in our pipeline to let the user choose a better one for the specific scene. Please note that even if the masks are not as consistent as the ground truth for the multiple views, our 2D mask optimization method can compensate for the errors. We also provide a multi-view version of Figure 8 of the paper in Figure \ref{fig:mask_mul}.

\section{VGG and LSeg Features Property}
As shown in Figure \ref{fig:vgg_lseg} of the paper, if we only use VGG features, the feature-matching algorithm produces some artifacts on the horse. If we only use LSeg features instead, the results become noisy. There are two possible reasons. Firstly, LSeg features are predicted pixel-wise and down-sampled to the same resolution as the VGG features. Since the optimization is performed on VGG distance with a lower resolution, LSeg-based feature matching will create the aliasing result. Secondly, LSeg cannot capture textural and color information well compared to VGG. Therefore, a combination of VGG and LSeg distance gives the best result by balancing the semantic and textural information. We can also observe that the smoothness of the style transfer result is increased as $\alpha$ increases, which confirms our first conjecture. Therefore, our combined distance gives additional controllability in terms of the balance of semantic-textural information and the smoothness of the final result.

\end{document}